\documentclass[3p,times]{elsarticle}

\usepackage{ecrc}

\volume{}

\firstpage{1}

\journalname{}

\runauth{}

\jid{procs}

\usepackage{amsfonts,amssymb,amsmath,amsthm}
\usepackage{multicol}
\usepackage{graphicx}
\usepackage{epsfig}
\usepackage{subfigure}
\usepackage{enumerate}
\usepackage{color}
\usepackage{comment}
\usepackage{tikz} 
\usepackage[caption=false]{subfig}
\usepackage{caption}
\usepackage{lscape}
\usepackage{booktabs}
\usepackage{multicol}
\usepackage{xcolor,colortbl}
\definecolor{lavender}{rgb}{0.9, 0.9, 0.98}
\usepackage{url}

\usepackage{lineno}

\usepackage[figuresright]{rotating}

\newtheorem{thm}{Theorem}[section] 
\newtheorem{cor}[thm]{Corollary}

\newtheorem{rem}{Remark}

\numberwithin{equation}{section}

\begin{document}

\begin{frontmatter}


\title{Title\tnoteref{label1}}

\dochead{}
%% e.g. \dochead{17th International Conference on Dynamical Processes in Excited States of Solids}

\title{%(Compact)
Hermite coordinate interpolation kernels: application to image zooming   
}


\author[label1]{Konstantinos K. Delibasis\corref{cor1}}
\ead[label1]{kdelibasis@gmail.com}
\cortext[cor1]{Corresponding author. Tel.: (+30) 22310 66908.}
\address[label1]{Department of Computer Science and Biomedical Informatics, University of Thessaly, 2-4 Papasiopoulou str., P.O. 35131 Lamia, Greece}

\author[label3]{Iro Oikonomou}
\ead[label3]{iro.oikonomou99@gmail.com}

\address[label3]{Department of Informatics and Telecommunications, National and Kapodistrian University of Athens,
Panepistimioupolis, Ilisia 157 84, Athens Greece }

\author[label2]{Aristides I. Kechriniotis}
\ead[label2]{arisk7@gmail.com}
\address[label2]{Department of Physics, University of Thessaly, 3rd Km Old National Road Lamia–Athens
35100, Lamia
Greece}

\author[label4]{Georgios N. Tsigaridas}
\address[label4]{Department of Physics
School of Applied Mathematical and Physical Sciences,
National Technical University of Athens,
Zografou Campus
GR-15780 Zografou, Athens
Greece}
\ead[label4]{gtsig@mail.ntua.gr}

\begin{abstract}
A number of basic image processing tasks, such as any geometric transformation require interpolation at subpixel image values. In this work we utilize the multidimensional coordinate Hermite spline interpolation defined on non-equal spaced, rectilinear grids and apply it to a very common image processing task, image zooming. Since Hermite interpolation utilizes function values, as well as partial derivative values, it is natural to apply it to image processing tasks as a special case of equi-spaced grid, using numerical approximations of the image partial derivatives at each pixel. Furthermore, the task of image interpolation requires the calculation of image values at positions with nono-zero fractional part. Thus, any spline interpolation can be written as convolution with an appropriate kernel. In this context we generate the Hermite kernels according to the derived $n-$dimensional interpolant of Theorem 2 in \cite{DK23}. We show that despite the increased complexity of the interpolant, once the kernels are constructed, the Hermite spline interpolation can be applied to images as efficiently as any other less complicated method. Finally, we perform illustrative numerical examples to showcase the applicability and high accuracy of the proposed Hermite kernels for image zooming, compared to other interpolation methods, both traditional convolution-based, as well as employing deep learning, in terms of PSNR, as well as SSIM error metrics. The proposed Hermite spline kernels outperform all other methods in the majority of the test images, in experiments using many cascaded repetitions of the zoom operation. Interesting conclusions can be drawn considering all methods under comparison.
%we study the Hermite interpolation on $n$-dimensional non-equal spaced, rectilinear grids over a field $\Bbbk $ of characteristic zero, given the values of the function at each point of the grid and the partial derivatives up to a maximum degree. First, we prove the uniqueness of the interpolating polynomial, and we further obtain a compact closed form that uses a single summation, irrespective of the dimensionality. We provide the remainder of the interpolation in integral form; moreover, we derive the ideal of the interpolation and express  the interpolation remainder using only polynomial divisions, in the case of interpolating a polynomial function. The arithmetic complexity of the derived closed formula compares favourably with the only alternative closed form for the $n$-dimensional classical Hermite interpolation \cite{fractmult}. .

\end{abstract}

\begin{keyword}
polynomial image interpolation \sep multivariate Hermite \sep Hermite spline kernels \sep image zoom

%\PACS 03.65.Pm \sep 03.50.De \sep 41.20.-q


\end{keyword}

\end{frontmatter}

% \linenumbers

%Introduction
\section{Introduction}
Any image geometric transformation requires image values at new, non-integer positions. The classic approach to this task is to employ an interpolation method using the given image values at neighboring pixels, as support points. Since any image consists of hundreds or even thousands of lines and columns, every image interpolation method is implemented as splines, piece-wise continuous polynomial functions defined over a small number of support points (image pixels), to maintain a low degree.

In this work, we utilize the definition of splines based on Hermite polynomials on $n$-dimensional grids, \cite{DK23}, to define $n$-dimensional Hermite spline convolution kernels. Hermite interpolation has been proposed for signal and image applications \cite{DK12}, \cite{DK14} achieving very competitive results, however these implementations were based on Theorems specifically for 1D and 2D, not easily generalizable to higher dimensions, whereas they did not consider spline construction, thus their execution was slow. The construction of Hermite kernels greatly accelerates the execution of Hermite interpolation, which otherwise would be of increased arithmetic complexity. 
In general, for any interpolation technique using kernel $K(\mathbf{x_0})$, the value of an ($n$-dimensional) image $I$ at a non-integer location $\mathbf{x_0}$ (assuming integer pixel coordinates) is obtained as
$I(\mathbf{x_0})=\sum_{\mathbf{a}}I(\mathbf{a})K(\mathbf{x_0}-\mathbf{a})$, which is the discrete convolutional summation.

Let us consider the case of 2D image scaling by integer factor $s$, hereforth referred as zooming. In this very common geometric transformation, the fractional parts of the unknown image positions are constant. I.e., in the special case of $s=2$, for each given image pixel $(i,j)=(x,y)$, the new image values have to be computed at three new locations: $(x+0.5, y), (x,y+0.5),(x+0.5,y+0.5)$. Since any interpolation kernels depends on the fractional part of the required point, interpolation is most efficiently implemented by pre-constructing three different kernels $K^H_{10},K^H_{01},K^H_{11}$ and convolving the image with each one to generate the values at the intermediate points to the right, south and south-east of each of the initial image pixels, respectively. Figure \ref{figszoom_x2} depicts this process graphically using green squares for the given pixels and red, blue and yellow circles for the east, south and south-east pixels, respectively. This is a generic approach, applicable to any convolution-based method and it is considered computationally very efficient, since image convolution operations are implemented in highly parallelizable manner.

The rest of the paper is structured as follows. The $n$-dimensional Hermite spline is presented in Theorem \ref{mainform}. The generic Hermite kernel is constructed in section 4. The specific task of x2 image zoom is expressed as image interpolation task that requires only 3 kernels (by any convolution-based image interpolation method, one of them being the proposed $n-$ dimensional Hermite spline).

We experiment with different settings of the Hermite splines, namely the support point stencil $3 \times 5$ and $3 \times 3$ pixels, as well as the order of the image partial derivatives up to 2nd and 3rd order for each spatial dimension. We also explore the image derivative approximation and we include the Hermite kernel and the necessary derivative kernels into a single kernel, which further accelerates the execution and results in a very simple implementation. We also construct the corresponding kernels from other state-of-the-art interpolation methods, including the generalized convolution ones (maximal order minimal support- OMOMS- and b-splines), and we further include a few established super-resolution methods based on Deep learning. Comparative results are finally presented for real world image datasets.

\section{The proposed Hermite interpolation on $n$-$D$ rectilinear grids}

\subsection{Introduction and Notations of Hermite interpolation}
\label{intro}
\noindent
This work focuses on multivariate classical Hermite interpolation with support points arranged on an n-dimensional non-equally spaced rectilinear grid ($nD$ grid), given the value of a function, as well as its derivatives up to an arbitrary maximum order, defined independently for each point and each dimension.

\noindent Let $A$ be a set, $\left\vert A\right\vert $ the cardinality of $A$, and $A^{n}:=\underset{n-times}{A\times \dots\times A}$. Given the sets $A_{1},...,A_{n}$, then $\mathbf{A:=}A_{1}\times
...\times A_{n}$. Further, the element $\left( a_{1},...a_{n}\right) \in 
\mathbf{A}$ will be denoted by $\mathbf{a}$. Let 
 $\mathbf{0=}\left( 0,0,...,0\right) $, \ $\mathbf{1=}\left(
1,1,...,1\right)$ be the zero vector and ones vector,  respectively.
Thus, points $\mathbf{a} = \left( a_{1},...a_{n}\right)$ are arranged on a non-regular N-dimensional grid $\mathbf{A}$. Let $\Bbbk $ be a field of characteristic zero. For $\mathbf{a\in }\Bbbk ^{n}$, and $\mathbf{m\in}\mathbb{N}_{0}^{n}$ we denote $\mathbf{a}^{\mathbf{m}}:=\prod_{i=1}^{n}a_{i}^{m_{i}}$. 
Let $\mathbf{k} = \left( k_{1},...k_{n}\right)$ be an $n$-dimensional vector of non-negative integers, holding the order of partial derivatives of the interpolating polynomial with respect to each variable.

In $\mathbb{N}_{0}^{n}$ we define the relation "$\leq "$ as follows: $\mathbf{k}\leq 
\mathbf{m}$ if and only if $k_{i}\leq m_{i}$, for every $i=1,...,n.$ Clearly 
$\left( \mathbb{N}_{0}^{n},\leq \right) $ is a poset ($\mathbb{N}_{0}^{n}$ is partially ordered). If $\mathbf{k\leq m}$ and $\mathbf{k\neq m}$, then $\left[ \mathbf{k,m}\right] :=\left\{ \mathbf{l\in }\mathbb{N}_{0}^{n}:\mathbf{k\leq l\leq m}\right\} =\left[ k_{1},m_{1}\right] \times
\dots \times \left[ k_{n},m_{n}\right] $, and is valid $\left\vert \left[ 
\mathbf{k,m}\right] \right\vert =\prod_{i=1}^{n}\left( m_{i}-k_{i}+1\right) $.

Given the finite subsets $A_{i}$, $i=1,\dots ,n$ of the field $\Bbbk $,
and the multiplicity functions $\nu _{i}:A_{i}\rightarrow 
\mathbb{N},~i=1,\dots ,n$. Then for $i\in \left\{ 1,\dots ,n\right\} $, and $a_i$ any element of $A_{i}$, we define

\begin{equation}\label{Hmikra}
%H_{\left( i,a\right) }\left( x_{i}\right) 
H_{a_i}\left( x_i\right) :=\prod_{\substack{ c\in A_{i}  \\ %
c\neq a_i}}\left( \frac{x_{i}-c}{a_i-c}\right) ^{\nu _{i}\left( a\right) }\in \Bbbk
\left[ x_{i}\right] .
\end{equation}

Let $\nu :\mathbf{A}\rightarrow \mathbb{N}^{n}$ be the generalized multiplicity function given by $\nu \left( 
\mathbf{a}\right) :=\left( \nu _{1}\left( a_{1}\right) ,\dots,\nu _{n}\left(
a_{n}\right) \right) $. For $\mathbf{a\in A}$ and $\mathbf{k\in }\left[ 
\mathbf{0,}\nu \left( \mathbf{a}\right) -\mathbf{1}\right] $ we define
\begin{equation}\label{Hmegala}
H_{\left( \mathbf{a,k}\right) }\left( x_{1},\dots, x_{n}\right) :=\prod_{i=1}^{n}%
\frac{\left( x_{i}-a_{i}\right) ^{k_{i}} H_{a_i}\left( x_i\right) }{k_{i}!}\in \Bbbk \left[ x_{1},\dots ,x_{n}\right] .
\end{equation}

We define the partial derivative operator acting on $f$ as
$$
\partial ^{\mathbf{k}}:=\prod_{i=1}^{n}\partial _{i}^{k_{i}},\; \partial _{%
\mathbf{a}}^{\mathbf{k}}f\left( \mathbf{x}\right) :=\partial ^{\mathbf{k}%
}f\left( \mathbf{x}\right) \left\vert _{\mathbf{x=a}}\right. ,\;\text{where } \partial _{i}^{k}:=\frac{\partial ^{k}}{\partial x_{i}^{k}}.$$

The article is organized as follows. Having defined some necessary notations, we provide in the next section the...

\label{mainpart}

\begin{rem}\label{rem1}
{\rm
It is easy to verify that 
\begin{eqnarray*}
\left\vert \left\{ \mathbf{a}^{\mathbf{k}}:\text{ }\mathbf{a\in A}\text{, }%
\mathbf{k\in }\left[ \mathbf{0,}\nu \left( \mathbf{a}\right) -\mathbf{1}%
\right] \right\} \right\vert =\sum_{\mathbf{a\in A}}\sum_{\mathbf{k\in }%
\left[ \mathbf{0,}\nu \left( \mathbf{a}\right) -\mathbf{1}\right] }1= 
\sum_{\mathbf{a\in A}}\prod_{i=1}^{n}\nu _{i}\left( a_{i}\right)
=\prod_{i=1}^{n}\sum_{a\in A_{i}}\nu _{i}\left( a\right) .
\end{eqnarray*}
}
\end{rem}

The set of partial derivative operators has equal cardinality: $$\left\vert \left\{ \partial _{%
\mathbf{a}}^{\mathbf{m}},~\mathbf{a\in A},~\mathbf{m\in }\left[ \mathbf{0,}%
\nu \left( \mathbf{a}\right) -\mathbf{1}\right] \right\} \right\vert
=\prod_{i=1}^{n}\sum_{a_{i}\in A_{i}}\nu _{i}\left( a\right) .$$

\begin{rem}\label{rem4}
{\rm
For $\mathbf{a,b\in A}$, and $\mathbf{k,m\in }\left[ \mathbf{0,}\nu \left( 
\mathbf{a}\right) -\mathbf{1}\right] $ by using the Leibniz derivative rule
we easily get:
\begin{equation*}
\partial _{\mathbf{a}}^{\mathbf{k}}H_{\left( \mathbf{b,m}\right) }=\left\{ 
\begin{array}{cccccc}
0, & \text{if} & \mathbf{a\neq b} &  &  &  \\ 
0, & \text{if} & \mathbf{a=b} & \text{and} & \mathbf{k<m} &  \\ 
0, & \text{if} & \mathbf{a=b} & \text{and} & \mathbf{k,m} & \text{are
incomparable} \\ 
1, & \text{if} & \mathbf{a=b} & \text{and} & \mathbf{m=k} & 
\end{array}%
\right. .
\end{equation*}
}
\end{rem}
%}

We now present the Theorem for the uniqueness of the multivariate Hermite interpolating polynomial.
\begin{thm}\label{thm1}
Given the elements $t_{\mathbf{a}}^{\mathbf{k}}\in \Bbbk \,,~\mathbf{a\in A}%
,~\mathbf{k\in }\left[ \mathbf{0,}\nu \left( \mathbf{a}\right) -\mathbf{1}%
\right] $, there exists a unique $f\in V\left( \mathbf{A,}\nu \right) $
such that $\partial _{\mathbf{a}}^{\mathbf{m}}f=t_{\mathbf{a}}^{\mathbf{m}},~%
\mathbf{a\in A},~\mathbf{m\in }\left[ \mathbf{0,}\nu \left( \mathbf{a}%
\right) -\mathbf{1}\right]$.
\end{thm}

We will derive an expression for the interpolating polynomial $\ f$
in Theorem \ref{thm1}. First we will use the degree reverse lexicographic order,
which will be denoted by $\prec $ . More specifically,$(k_1,\dots, k_n) \prec (l_1,\dots l_n)$, if either of the following holds:
\begin{itemize}
    \item[(i)] $k_1 +\dots + k_n < l_1 +\dots +l_n$, or
    \item[(ii)] $k_1 +\dots +k_n = l_1 +\dots+l_n$ and $k_i > l_i$ for the largest $i$ for which $k_i \neq l_i$.
\end{itemize}
For example, the reverse lexicographic order  of the elements in $\left[ \mathbf{0,}\nu \left( \mathbf{a}\right) -\mathbf{1}\right] $ is
\begin{eqnarray}\label{lexord}
1_{\mathbf{a}} :&=&\left( 0,\dots,0\right) \prec 2_{\mathbf{a}}:=\left(
0,\dots,0,1\right) \prec 3_{\mathbf{a}}:=\left( 0,\dots,0,1,0\right) \prec
\dots\prec \left( n+1\right) _{\mathbf{a}}:=\left( 1,0,\dots,0\right)\nonumber \\
&\prec &\left( n+2\right) _{\mathbf{a}}:=\left( 0,\dots,0,2\right) \prec
\dots\prec ({\left\vert \nu \left( \mathbf{a}\right) \right\vert -1})_{\mathbf{a}}:=\left(
\nu _{1}\left( a_{1}\right) -1,...,\nu _{n}\left( a_{n}\right) -1\right) .
\end{eqnarray}%
Note that from $\mathbf{m\leq n}$ follows $\mathbf{m\preceq n}$, and from $%
\mathbf{m<n}$ follows $\mathbf{m\prec n}$. That means $\mathbf{\preceq }$ is
a linear extension of $\mathbf{\leq }$.

The following Theorem provides the closed form of the interpolating polynomial.
\begin{thm}\label{finterpform}
The formula of the interpolating
polynomial $\ f$ is the following
\begin{equation} \label{mainform}
f(\mathbf{x}):=\sum\limits_{\mathbf{a\in A}}H_{%
\mathbf{a}}^T \Lambda _{\mathbf{a}}^{-1}T_{\mathbf{a}}=
\sum\limits_{\mathbf{a\in A}}\sum_{i=1}^{\left\vert \nu \left( 
\mathbf{a}\right) \right\vert -1 } H_{%
\mathbf{a}}^T  \left( I_{\left\vert \nu \left( \mathbf{a}%
\right) \right\vert }-\Lambda _{\mathbf{a}}\right)^i T_{\mathbf{a}},
\end{equation}%
where $H_{\mathbf{a}}=H_{\mathbf{a}}(\mathbf{x})$,
\begin{eqnarray}\label{La}
\Lambda _{\mathbf{a}} &=&\left[ 
\begin{array}{ccccc}
1 & 0 & \cdots &  0 & 0 \\ 
%\varepsilon _{2,1}
\varepsilon _{2,1}\partial _{\mathbf{a}}^{2_{\mathbf{a}}}H_{\left( \mathbf{a,%
}1_{\mathbf{a}}\right) } & 1 & \cdots & 0 & 0 \\ 
\vdots & \vdots & \ddots  & \vdots & \vdots \\ 
%\varepsilon _{\left\vert \nu \left( \mathbf{a}\right) \right\vert-1,1}
\varepsilon _{\left\vert \nu \left( \mathbf{a}\right) \right\vert-1
,1} \partial _{\mathbf{a}}^{\left( \left\vert \nu \left( \mathbf{a}\right)
\right\vert -1\right) _{\mathbf{a}}}H_{\left( \mathbf{a,}1_{\mathbf{a}%
}\right) } 
& 
\varepsilon_{\left\vert \nu \left( \mathbf{a}\right)\right\vert -1,2}
\partial _{\mathbf{a}}^{\left( \left\vert \nu \left( 
\mathbf{a}\right) \right\vert -1\right) _{\mathbf{a}}}H_{\left( \mathbf{a,}%
2_{\mathbf{a}}\right) } 
&\cdots & 1 & 0 \\ 
\varepsilon _{\left\vert \nu \left( \mathbf{a}\right) \right\vert,1}
\partial _{\mathbf{a}}^{\left\vert \nu \left( \mathbf{a}\right)
\right\vert _{\mathbf{a}}}H_{\left( \mathbf{a,}1_{\mathbf{a}}\right) } 
& 
\varepsilon _{\left\vert \nu \left( \mathbf{a}\right) \right\vert,2}
\partial _{\mathbf{a}}^{\left\vert \nu \left( \mathbf{a}\right)
\right\vert _{\mathbf{a}}}H_{\left( \mathbf{a,}2_{\mathbf{a}}\right) } & 
\cdots & 
\varepsilon _{\left\vert \nu \left( \mathbf{a}\right) \right\vert,\left\vert \nu \left( \mathbf{a}\right) \right\vert -1}
\partial _{\mathbf{a}%
}^{\left\vert \nu \left( \mathbf{a}\right) \right\vert _{\mathbf{a}%
}}H_{\left( \mathbf{a,}\left( \left\vert \nu \left( \mathbf{a}\right)
\right\vert -1\right) _{\mathbf{a}}\right) } & 1%
\end{array}%
\right] , \\
x_{\mathbf{a}} &=&\left[ 
\begin{array}{c}
x_{\mathbf{a}}^{1_{\mathbf{a}}} \\ 
\vdots
\\
x_{\mathbf{a}}^{\left\vert \nu \left( \mathbf{a}\right) \right\vert }%
\end{array}%
\right] \text{, }T_{\mathbf{a}}=\left[ 
\begin{array}{c}
t_{\mathbf{a}}^{1_{\mathbf{a}}} \\ 
\vdots
\\
t_{\mathbf{a}}^{\left\vert \nu \left( \mathbf{a}\right) \right\vert }%
\end{array}%
\right] ,H_{\mathbf{a}}=\left[ 
\begin{array}{c}
H_{\left( \mathbf{a,}1_{\mathbf{a}}\right) } \\ 
\vdots
\\
H_{\left( _{\mathbf{a,}\left\vert \nu \left( \mathbf{a}\right) \right\vert
}\right) }%
\end{array}%
\right] 
,\; \lvert \nu \left( \mathbf{a}\right) \rvert =\prod_{i=1}^{n}\nu _{i}\left( a_{i}\right).\nonumber 
\end{eqnarray}
\end{thm}

In our recent work \cite{DK23} we also established the $n$-dimensional Hermite spline for even, as well as odd number of local support points, which is also referred as stencil, and proved its continuity. Contrary to the most popular convolution-based implementations, the multi-dimensional Hermite interpolant is a non-separable polynomial.

%\section{Hermite Spline interpolation}
%\input{splinsection}

\section{Hermite spline interpolation Kernel}
\label{constructkernel}
\subsection{Hermite spline interpolation as Convolution and Kernel definition}
By considering the application of Theorem \ref{finterpform} to a local image neighborhood as a spline, it is easy to observe that the summation in Eq. \ref{mainform} can be thought of as $n+1$-dimensional convolution of image data $T_{\mathbf{a}}$ with the remaining quantity $ H^{T}_{\mathbf{a}} \Lambda^{-1}_{\mathbf{a}}$. Please note that the extra dimension when dealing with $n$-dimensional data comes from the use of partial derivatives of the data. The length of this dimension is equal to the cardinality of the reverse lexicographic vectors $\displaystyle c=\prod_{i=1}^{n}{\nu_i}=\nu_0^n,$, where the value of the multiplicity $\nu(\mathbf{a})=(\nu_1, \ldots \nu_n)=\nu_0$ is constant for all support points.
Let us further define the notation for the image derivatives for any derivative vector $\mathbf{k}$, $I_{\mathbf{k}}=I_{k_1,\ldots k_n}=\frac{\partial^{k_1+\ldots k_n}I}{\partial{x_{1}^{k_1}} \ldots \partial{x_{n}^{k_n}}}$.

Expressing the interpolation as convolution has the profound advantage of very fast and highly parallelizable execution, especially in parallel architectures such as Graphics Processing Units (GPUs), assuming of course that the convolution Kernel is constant. As with any convolution -based spline interpolation method, the kernel depend on the fractional part of the coordinates where the image values is to be interpolated, in our case $\mathbf{x_0}- \lfloor \mathbf{x_0}\rfloor$.
As in any image or signal interpolation task, the number of support points is equal in each dimension $|A_1|=,\ldots,|A_n|=N$. In certain cases, the fractional part of the required points $ \text{fractional}(x_0)=x_0- \lfloor x_0 \rfloor$ is constant, or it may only obtain very few specific values. For instance, when generating an intermediate frame in a video sequence, or generating an intermediate slice in a 3D volumetric image, then the fractional part of all required points is constant: $fractional(x_0)=[0,0,0.5]$. In the case of zooming into a 2D image by a factor of 2, the fractional parts of the required points are the following: $[0.5,0]$, $[0,0.5]$, and $[0.5,0.5]$.
It is evident that very popular tasks of images and multidimensional signal processing end up as interpolations at points with constant fractional parts. 

In such cases, most of the calculations of the proposed Hermite interpolation are repeated. Equivalently, it is possible to express the calculation of the interpolated value as a convolution by an appropriate \textit{constant} kernel. This kernel depends only on the number of support points $N$ per dimension, the value of the multiplicity function $\mathbf{\nu}$ and the fractional part of $x_0$, thus it can be precalculated and used in any image.

In the case of a constant functional part of the coordinates of the required point $x_0$, $H_{\mathbf{a}} \in R^{c \times 1} $ is a column with the values for each support point $\mathbf{a}.$ 
Thus, the size of $H_{\mathbf{a}}$ for all support points is $|A_1| \times \ldots \times |A_n| \times c$. The calculation of the values of $H_{\mathbf{a}}$ is performed only once: given the local support points, the coefficients of the univariate polynomials $H_{a_i}(x_i)$ are stored in an array of dimensions $|A_1| \times \ldots \times |A_n| \times c \times n \times \deg_i$. An efficient computational implementation of Hermite splines lies in the construction of convolution kernels using the formulas (\ref{La}). 
The lower triangular matrices are of dimension $N_1 \times \ldots \times N_n \times c \times c$ can be pre-calculated and inverted $\Lambda_{\mathbf{a}}^{-1}.$ 

According to the definition of the convolution operator, the kernel needs to be mirrored with respect to its center. Thus, we can construct a tensor $K$ of size $N_1 \times \ldots \times N_n \times c$ that will be used as a convolution kernel, as follows: 
$$K(i_1,\ldots ,i_n,:)=\underbrace{H_{(N_1-i_1+1,\ldots,N_n-i_n+1,:)}^{T}}_{1 \times c} \underbrace{\Lambda^{-1}_{(N_1-i_1+1,\ldots,N_n-i_n+1,:,:)}}_{c \times c}.$$
In the above we used the tensor notation $X_{i,j,:}$ where : denotes the elements with any value of the corresponding index. The flipping of the indices except the last one is necessary to implement the calculations as a series of $n-$dimensional convolution operations (considering $c$ $n-$dimensional Hermite kernels. If Hermite kernels are considered as one $n+1-dimensional$, the convolution is implemented in $n+1$-dimensions flipping the last index as well. If the operation of correlation can also be performed efficiently, then there is no need for index flipping:
$$K(i_1,\ldots ,i_n,:)=H_{(i,\ldots,i_n,:)}^{T} \Lambda^{-1}_{(i_1 \ldots i_n,:,:)}.$$
Let us consider the 3D case: $n=3$, where the grid points are triplets: $\mathbf{a}=(i,j,k)$ and $|A_1|$.
$$K(i,j,k,:)=H_{(N_1-i+1,N_2-j+1,N_3-k+1,:)}^{T} \Lambda^{-1}_{(N_1-i+1,N_2-j+1,N_3-k+1,:,:)}$$

\begin{figure}
\begin{center}
    \includegraphics[scale=0.6]{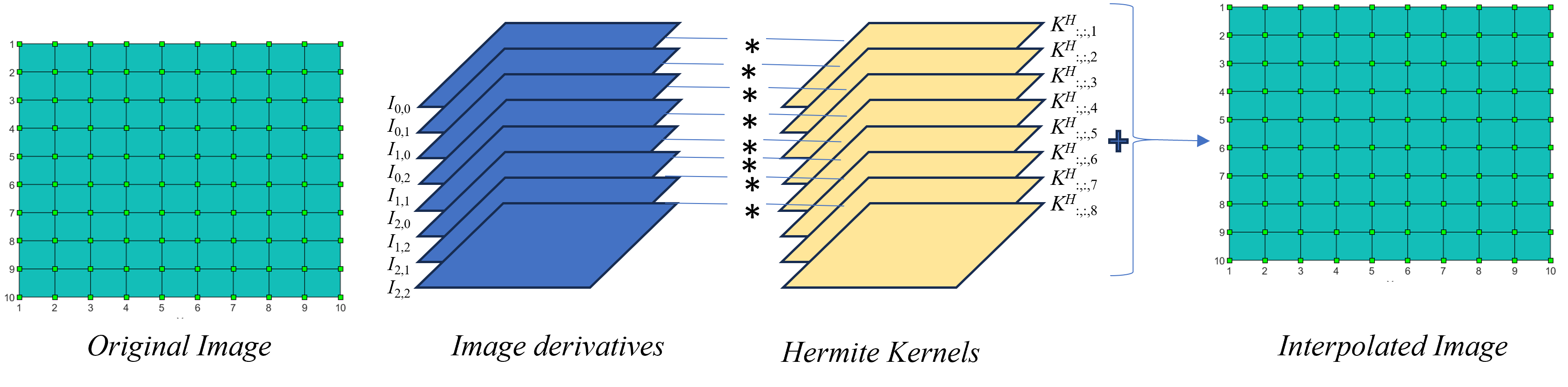}
    \caption{Diagram of convolution-based 2D Hermite spline interpolation with approximated image partial derivatives up to 1st order ($v=2$).}
    \label{2dcase}
\end{center}
\end{figure}

\begin{figure}[ht]
\begin{center}
    \includegraphics[scale=0.6]{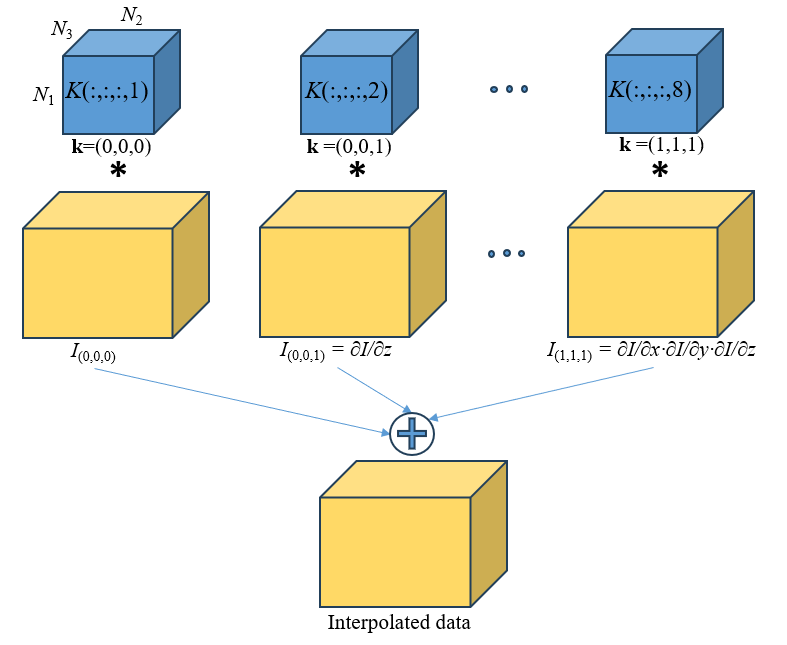}
    \caption{Diagram of convolution-based 3D Hermite spline interpolation with approximated image partial derivatives up to 1st order ($v=2$).}
    \label{3dcase}
\end{center}
\end{figure}

Given the complicated nature of Hermite splines, we have prepared and illustrated the procedural steps and the structural components required to apply Hermite kernels to 2D and 3D images, as depicted in Fig. \ref{2dcase} and Fig. \ref{3dcase}, respectively.
Therefore, having constructed the Hermite Kernels, we present the algorithmic steps and the data structures of applying Hermite kernels in Fig. \ref{2dcase} and Fig. \ref{3dcase} for 2D and 3D images respectively, using data derivatives. 
  
\begin{figure}[ht]
\begin{center}
    \includegraphics[scale=0.6]{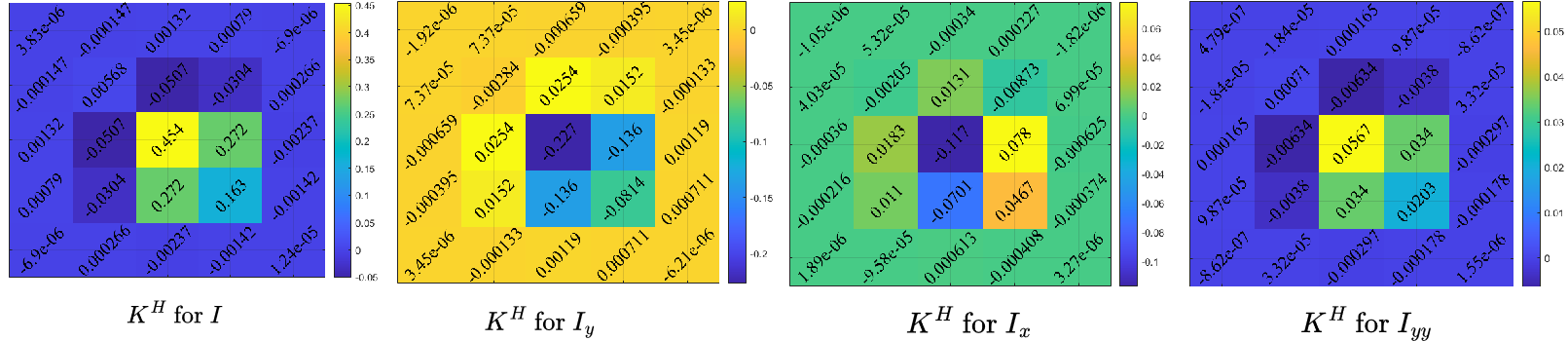}
    \caption{Numerical values for 2D Hermite spline interpolation for x2image zooming. The first 4 (out of 8) kernels are shown corresponding to approximated image partial derivatives up to 1st order ($v=2$).}
    \label{Herm_kernel_example}
\end{center}
\end{figure}

As an example, assuming that $\nu(\mathbf{a})=2$ for any grid point, equivalently all combinations of up to 1st order derivatives are concerned, consequently, the cardinality $c=4$. If i.e. we choose to use $N_1=N_2=5$ support points along each axis, the resulting Hermite kernel $K^H$ of size $5 \times 5 \times 4$. This is a 3rd order tensor which can also be treated as a series of 4 2D kernels, each of size $5 \times 5.$ In the case of $\nu(\mathbf{a})=3$ the kernel size becomes $5 \times 5 \times 8$. Fig. \ref{Herm_kernel_example} shows the first 4 slices of the corresponding Hermite kernel for interpolating images at (0.5, 0.5) with respect to each pixel.  

\begin{figure}[ht]
\begin{center}
    \includegraphics[scale=0.62
    ]{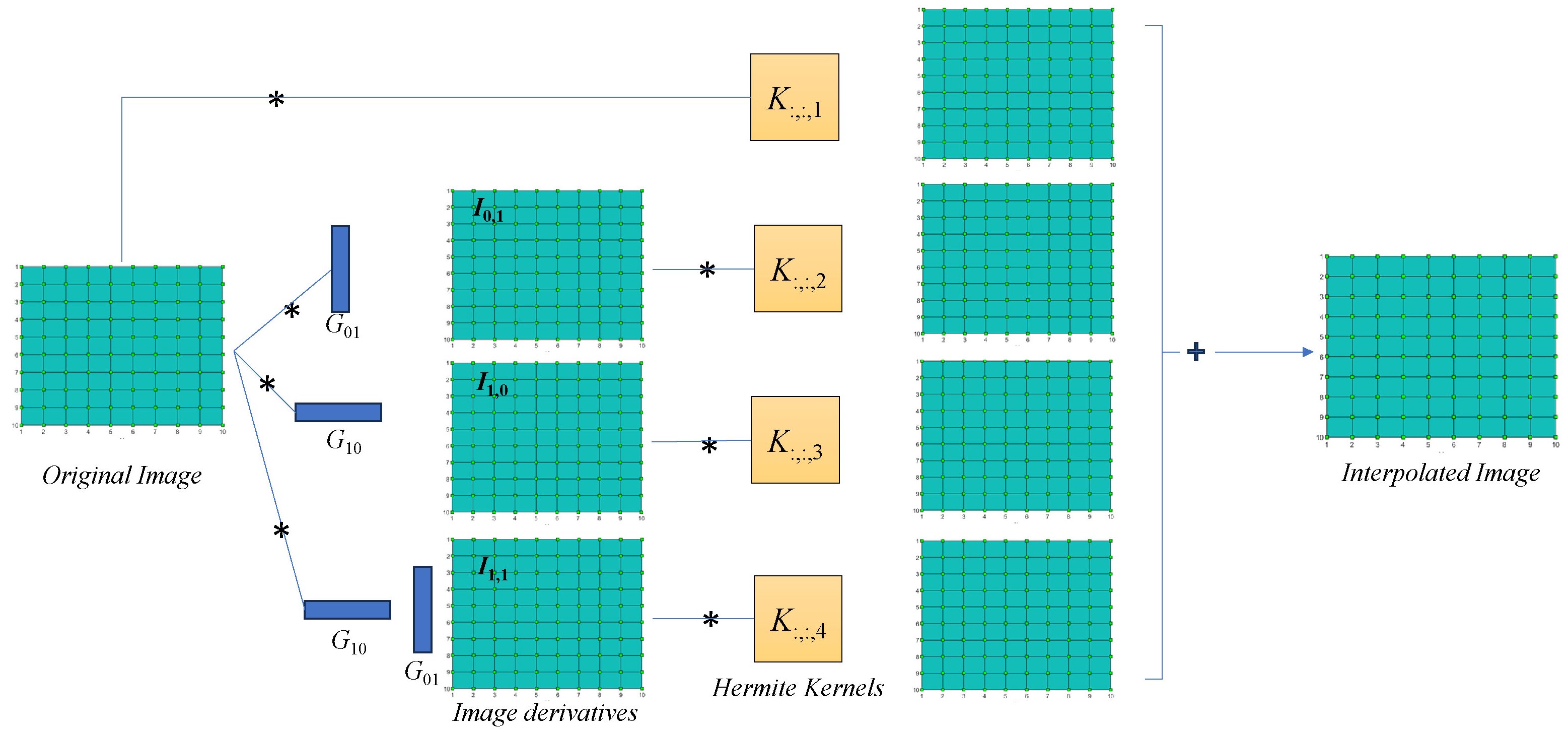}
    \caption{Diagram of applying Hermite spline interpolation on 2D image.}
\end{center}
\end{figure}

%%%%%%%%%%%%%%%%%%%%%%%%%%%%%
\subsection{Hermite interpolation Convolution for FIR-based numerically approximated data derivatives}
In the most frequent case of signal and image/video processing, the values of the derivatives are numerically approximated, using Finite impulse response (FIR) kernels, or Infinite impulse response (IIR), as described in \cite{delibasis2013derivatives}. Assuming the use of FIR kernels, it becomes possible to combine the FIR derivative kernels with the derived Hermite kernels into a single kernel with dimensionality equal to the dimensionality of the given data. 
More specifically, let us define the derivative kernel $G_\mathbf{k}$ that performs the image derivative approximation defined in vector $\mathbf{k}$, $ G_\mathbf{k} * I=I_{\mathbf{k}}=I_{k_1,\ldots k_n}=\frac{\partial^{k_1+\ldots k_n}I}{\partial{x_{1}^{k_1}} \ldots \partial{x_{n}^{k_n}}}$. For instance, $G_{02}$ is the kernel that approximate image derivatives of the 2nd order along the columns. Obviously, when $\mathbf{k}=\mathbf{0}$, then $G=[1]$ is the discrete Dirac function. The single $n-$dimensional Hermite Kernel that combines all partial derivatives of the given data is as follows:
$$\displaystyle \mathcal{K}=\sum_{j=1}^{c}G_{\mathbf{k_j}} \ast K(:,\dots,:,j),$$ where $\mathbf{k_j}$ refers to  the reverse lexicographic order  of the elements in $\left[ \mathbf{0,}\nu \left( \mathbf{a}\right) -\mathbf{1}\right]$  according to relation \ref{lexord}, $\mathbf{k_j}=j_{\mathbf{a}}$.

Let us consider a practical example of Hermite coordinate interpolation, applied to 2D images ($n=2$) and considering derivatives up to 1st degree $v(\mathbf{a})=(2,2)$ for any $\mathbf{a}$. The cardinality of the reverse lexicographic vectors is $c=4$, thus produces four (4) derivative vectors, according to \cite{delibasis2013derivatives}: $\mathbf{k_1}=(0,0)$, $\mathbf{k_2}=(0,1)$, $\mathbf{k_3}=(1,0)$, $\mathbf{k_4}=(1,1)$. The corresponding FIR derivative kernels, generated according to \cite{delibasis2013derivatives} are $G_{00},G_{01},G_{10} \text{ and } G_{11}$. It should be mentioned that $G_{00}=(1)$, $G_{01} \in \mathbb{R}^{q \times 1}$, $G_{10}=G_{01}^T$ and $G_{11}$ is approximated as $G_{10}*G_{01}$.

\begin{figure}[ht]
\begin{center}
    \includegraphics[scale=0.55]{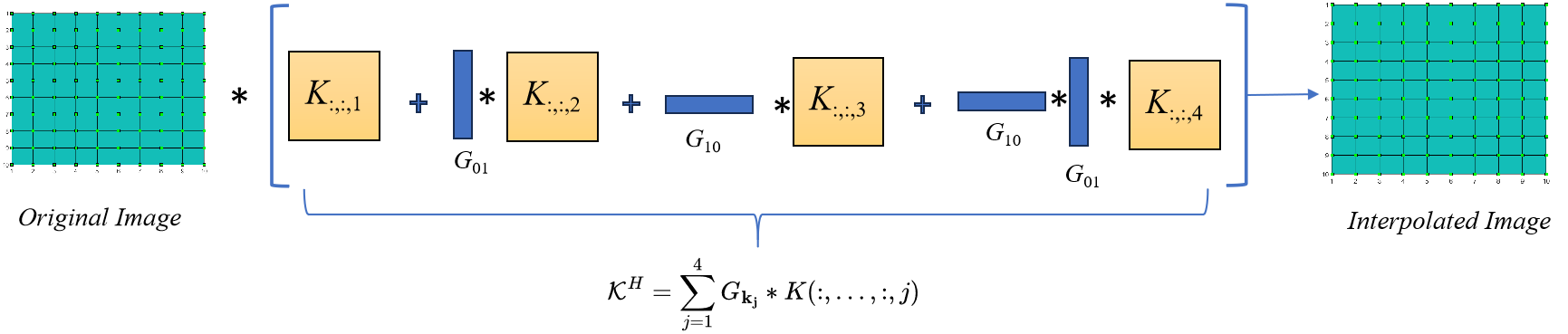}
    \caption{Principles of constructing a single 2D kernel that includes all 2D Hermite kernels, as well as the FIR differentiating kernels.}
    \label{Single_Herm_kernel}
\end{center}
\end{figure}

\subsection{Hermite interpolation Convolution for numerically approximated data derivatives using IIR}
\begin{figure}[!h]
\begin{center}
    \includegraphics[scale=0.5]{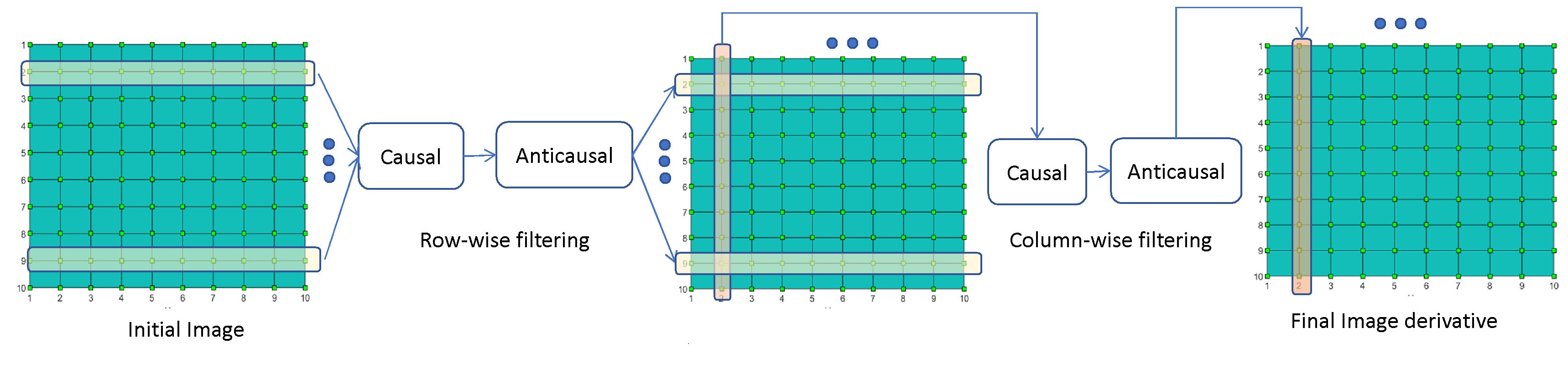}
    \caption{The steps of IIR-based implicit image derivative approximation.} \label{iir_image}
\end{center}
\end{figure}

A rather more complicated method to approximate signal and image derivatives is the implicit derivatives. As proposed by \cite{lele} for signals and implemented by \cite{belyaev1992implicit} for images, this method estimates derivatives implicitly:
\begin{equation*}
    Q_1 \ast f^{(1)}= R_1 \ast f
\end{equation*}
\begin{equation*}
    Q_2 \ast f^{(2)}= R_2 \ast f    
\end{equation*}
where $Q_1,Q_2$ are $5 \times 1$ symmetric kernels and $R_1$, $R_2$ are $7 \times 1$ antisymmetric and symmetric kernels respectively. The elements of each kernel is provided in \cite{lele}. In \cite{delibasis2013derivatives} the implementation of implicit derivatives is based on a cascade of two IIR filters: a causal and an anti-causal one with appropriate initial conditions and poles (roots of the denominator of the filters' transfer functions) that are pairs of reciprocals. The application to images is performed column-wise and then raw-wise in a very efficient manner. This process is shown in the diagram of Fig. \ref{iir_image}.
As it will be shown in the results, IIR-based image derivatives produce better Hermite interpolation compared to simpler FIR filters.

\subsection{Convolution-based implementation of 2d Image Super Resolution}
\begin{figure}[!ht]
     \centering
    \includegraphics[scale=0.5]{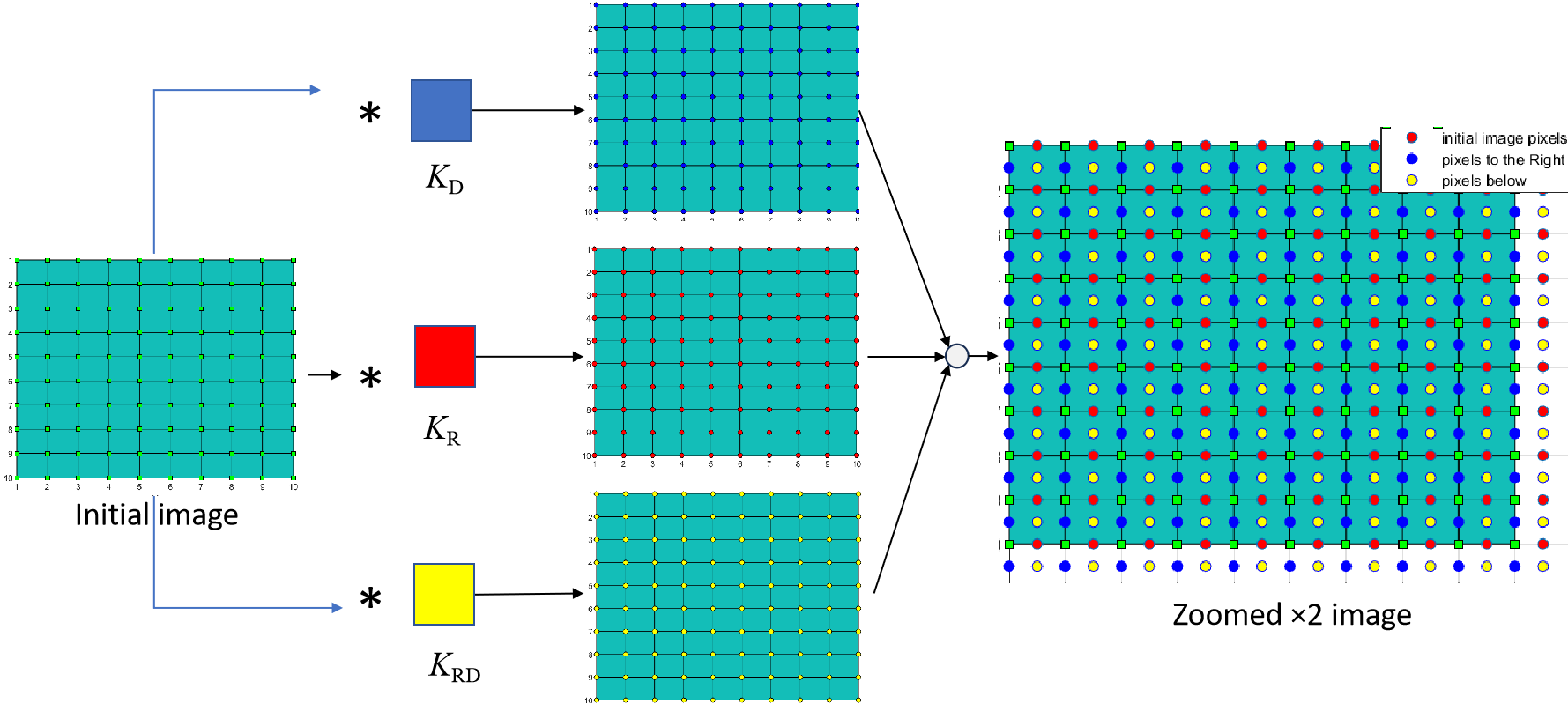}
         \caption{Using 3 constant kernels to implement 2D image zoom x2.}
         \label{figszoom_x2}
\end{figure}

Let us consider the case of 2D image scaling by integer factor $s$. In the special case of $s=2$, then for each given image pixel $(i,j)=(x,y)$, the new image values have to be computed at three new locations: $(x+0.5, y), (x,y+0.5),(x+0.5,y+0.5)$, or equivalently the value to the right of the current pixel, down and right-down of the current pixel, respectively.
This can be achieved most efficiently by constructing three different kernels:  $K^{HR},K^{HD},K^{HRD}$ and convolving the image with each one to generate the values at the corresponding pixels. Figure \ref{figszoom_x2} depicts this process graphically using green squares for the given pixels and red, blue and yellow circles for the east, south and south-east pixels, respectively. This is a generic approach, applicable to any convolution-based method and it is considered computationally very efficient, since image convolution operations are very efficiently implemented and highly parallelizable.

\subsection{Generic framework for the evaluation of interpolation methods}
\begin{figure}
    \centering
    \includegraphics[scale=0.65]{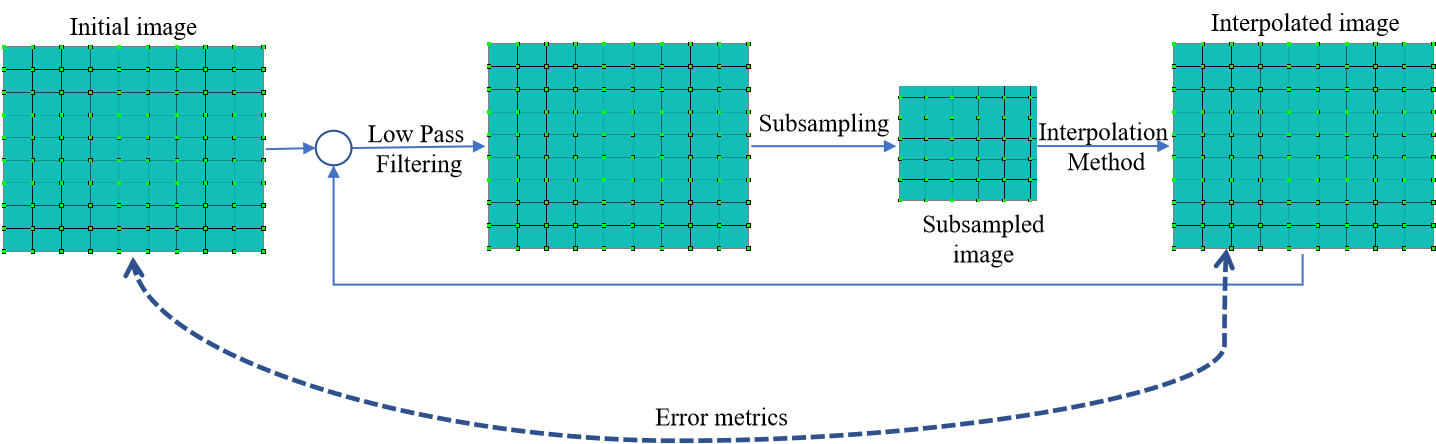}
    \caption{Scheme for quantitative evaluation of image zooming using multiple repetitions.}
    \label{gifzoom2dschm}
\end{figure}

We implemented a straightforward framework for the evaluation of interpolation methods on a large number of images, that has already been used in many works for this purpose. For any given image, the lines and columns (and slices in case of 3D images) are decimated, e.g. by removing the one every few of them (skip factor $s_k$). Obviously, the scale factor is given by $s=\frac{1}{s_k+1}$. Before this step, a low-pass (LP) smoothing filtering should be applied to alleviate the introduction of frequency aliasing. The normalized cut-off frequency of this filter should be chosen equal to the reciprocal of the scale factor $\omega_c=\pi\frac{1}{s}$ (assuming sampled frequency range $[-\pi,\pi].$ 
The interpolation methods are applied on the decimated image to generate an image of the original dimensions. An ideal interpolation would produce an image identical with the original. In order to exaggerate the differences between interpolation methods the previous steps are applied in a number of cascaded repetitions (using the interpolated image as the source for the next subsampling), thus generating an incremental error. 
The aforementioned steps are shown graphically in Fig. \ref{gifzoom2dschm}. 

%\section{Super resolution }
%\input{imagesupreds}

\section{Results}

\begin{figure}
    \centering
    \includegraphics[scale=0.7]{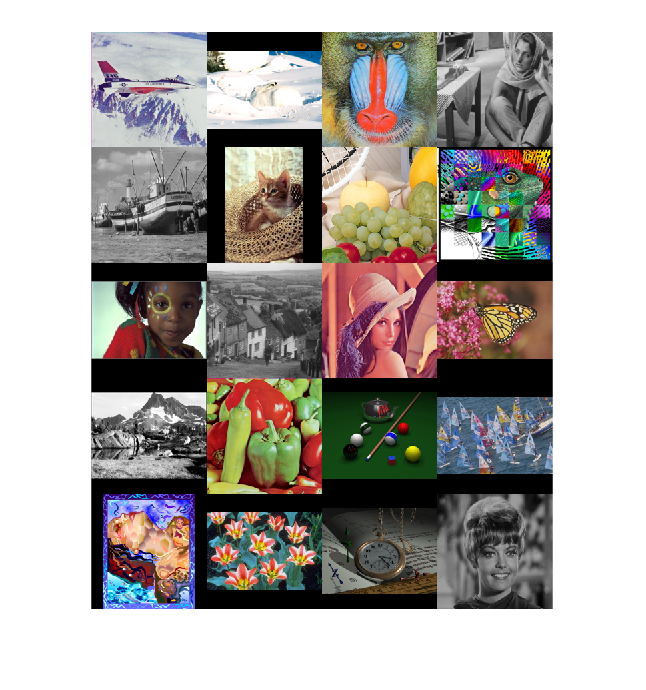}
    \caption{The dataset of the standard 20 images used for evaluation.}
    \label{figdata}
\end{figure}

The baseline for image interpolation is usually considered as the nearest neighbor one (expected to be the worst performer). Bilinear and bicubic interpolation are also convolution based methods that are also included in this work as base lines. In addition, we employed generalized convolution methods, which are considered state-of-the-art between the convolution-based methods. More specifically, the generalized convolution methods are using a pre-filtering step which consists of a number of pairs of causal and anti-causal filterings, each with reciprocal roots. We employed the best performing of these methods, namely b-spline \cite{unser1}, \cite{unser2} and maximal order minimal support (oMOM) up to the 5th degree (MOM4, MOM5) \cite{omoms1}. 
Finally, we included deep learning (DL) methods in the comparison. More specifically. we used the opencv implementation of three super-resolution neural networks with convolutional and deconvolutional layers, that generate zoomed versions of their input images, provided as already pretrained:
ESPCN \cite{espcn}
FSRCNN \cite{fsrcnn}
LapSRN \cite{lapsrn}.
In terms of the proposed Hermite spline kernels, we experimented with $N_1=N_2=5$ support points along each axis, and $\mathbf{\nu} =3$ resulting in Hermite kernels $K^H$ of size $5 \times 5 \times 9$ ($c=4$). The image partial derivatives up to 2nd order are approximated using 3, 5 and 7-point FIR filters, as well as IIR filters.

We utilized well established error metrics between the original image and the zoomed (after subsampling): PSNR, expressed in db (with greater values being better) and the Structural similarity index measure \cite{SSIM_Python} -SSIM. SSIM is a perception-based model that considers image degradation, including both luminance masking and contrast masking terms. The maximum value of 1 to indicate best interpolation quality. 

Results are presented in the case of 20 standard images, which have been benchmarks in image processing \cite{images_database}, shown in Fig. \ref{figdata}.

\begin{figure}[ht]
    \centering
    \includegraphics[scale=0.68]{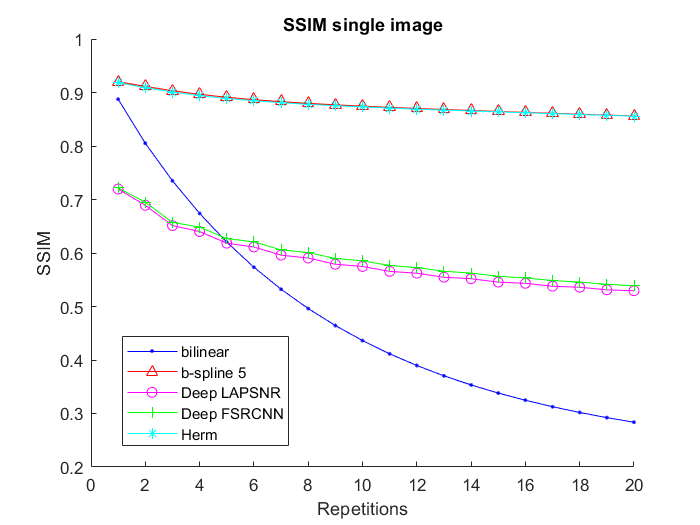}
    \caption{SSIM achieved by the methods in comparison for the Barbara image, using multiple repetitions of image zooming.}
    \label{SSIM_barbara}
\end{figure}

\begin{figure}
    \centering
    \includegraphics[scale=0.7]{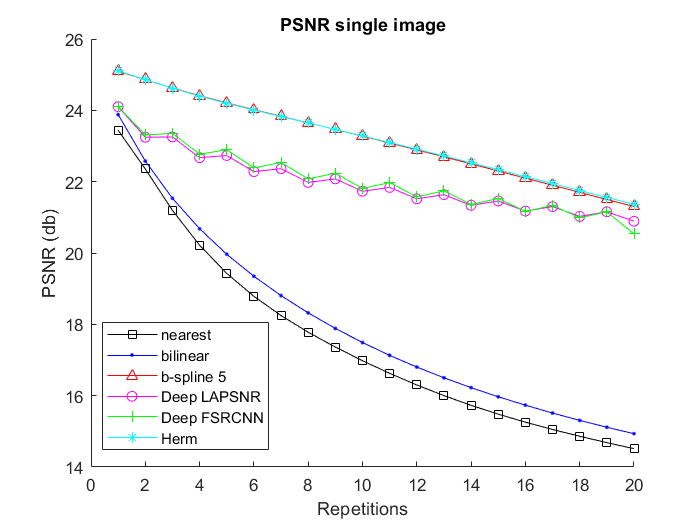}
    \caption{PSNR achieved by the methods in comparison for the barbara image using multiple repetitions of zooming.}
    \label{PSNR_barbara}
\end{figure}

\begin{figure}
    \centering
    \includegraphics[scale=0.7]{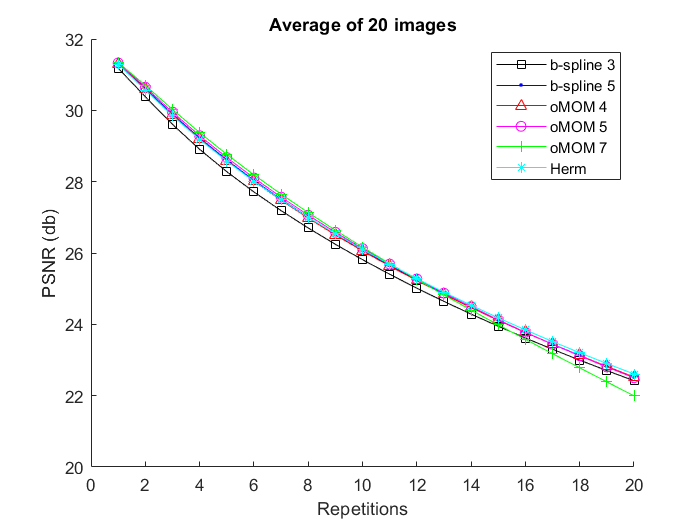}
    \caption{Average PSNR over all images for the best performing methods, using multiple repetitions of image zooming.}
    \label{PSNR_mean_all_img}
\end{figure}

\begin{figure}
    \centering
    \includegraphics[width=\textwidth,height=0.9\textheight]{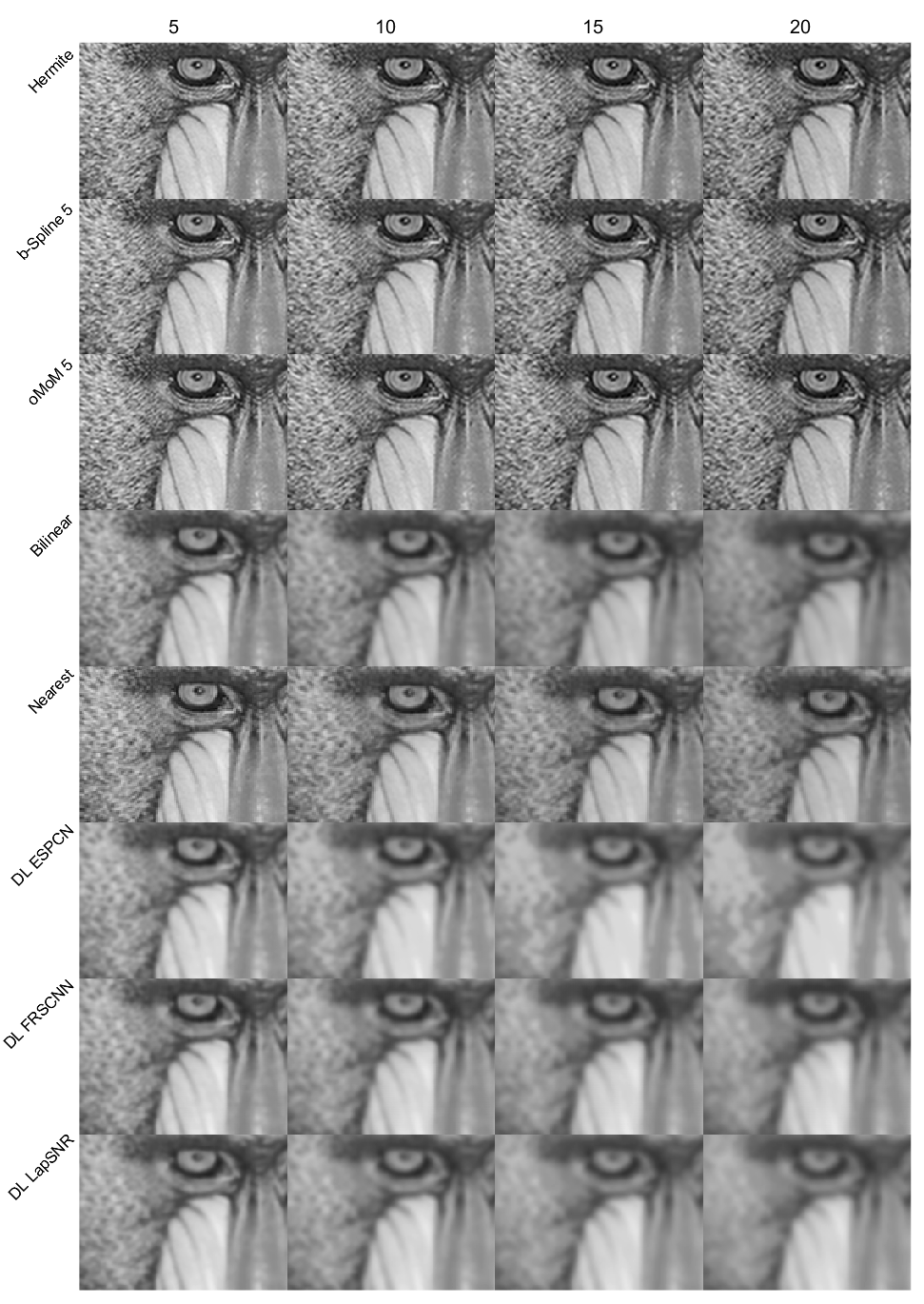}
    \caption{A cropped region from the 4th image (Baboon) interpolated by several methods, for  5, 10, 15 and 20 repetitions of the 2x zoom in and out operator.}
    \label{fig baboon_cropped_composite}
\end{figure}

\begin{figure}[!ht]
    \centering
    \includegraphics[width=\textwidth,height=0.9\textheight]{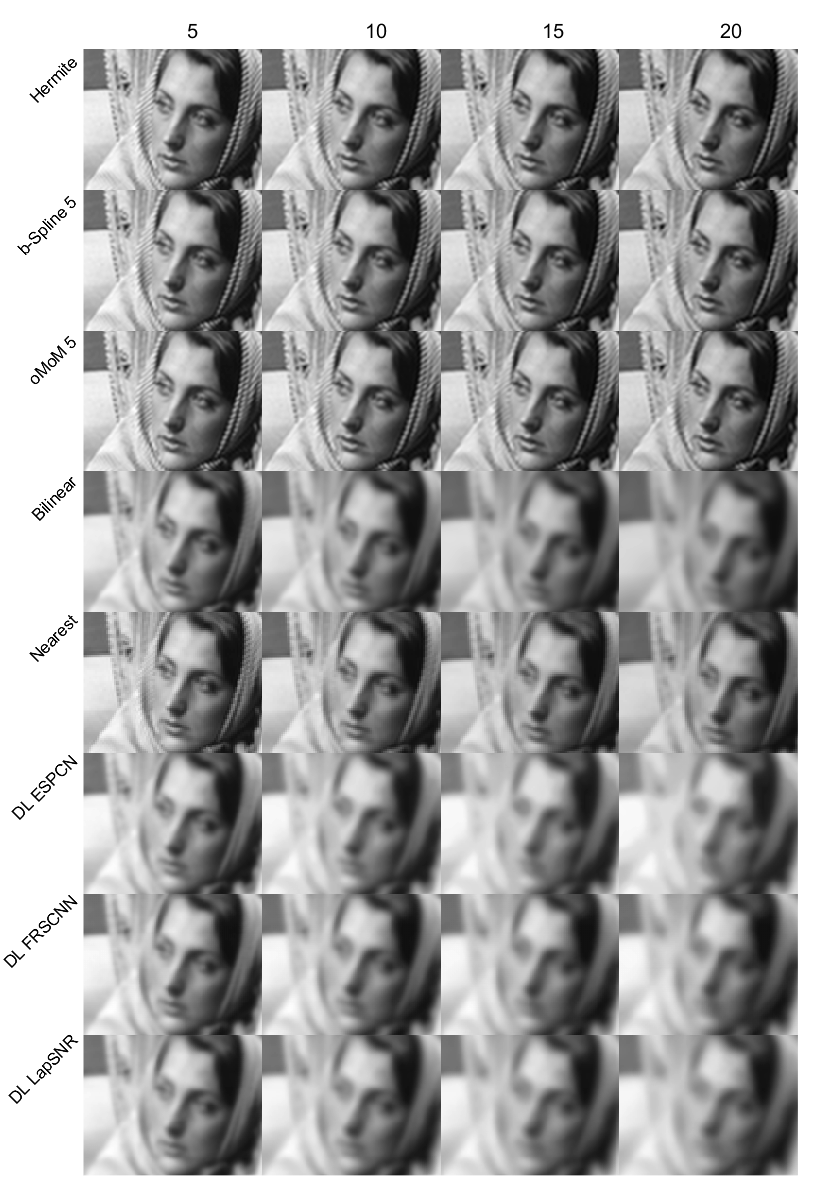}
    \caption{A cropped region from the 4th image (Barbara) interpolated by several methods, for  5, 10, 15 and 20 repetitions of the 2x zoom in and out operator.}
    \label{fig barbara_cropped_composite}
\end{figure}

Figure \ref{SSIM_barbara} shows the achieved SSIM by proposed Hermite with IIR image derivatives, the b-spline deg 5, two deep learning approaches, and the bilinear as baseline method for the Barbara image. In Figure \ref{SSIM_barbara}, while both the b-spline a deg 5 and the proposed Hermite method with IIR image derivatives exhibit superior performance, it's noteworthy that the Hermite method achieves a slight edge over the b-spline.
 The bilinear deteriorates very rapidly as the cascaded repetitions increase. The two deep learning methods also deteriorate significantly more than the best performers with increasing repetitions. A similar result is shown in Fig. \ref{PSNR_barbara} in terms of PSNR for the same image. In this case, the deep learning methods appear to deteriorate more slowly, seemingly converging with the best performers beyond the 20 repetitions. However, a visual inspection of the interpolated images in Figure \ref{fig baboon_cropped_composite} and \ref{fig barbara_cropped_composite} indicates that the blurring induced by the methods is very apparent. The high PSNR values are due to smaller errors in the flat image areas, whereas the fine image details are not preserved.
Figure \ref{PSNR_mean_all_img} shows average PSNR form all 20 images achieved by the proposed Hermite with FIR 7-point image derivatives and 5 more generalized convolution methods. The proposed Hermite spline method marginally outperforms the rest of the methods.

Detailed numeric results for each image and each of the methods are shown in Table \ref{psnr_table}
and Table \ref{SSIM_table}, after the 20 repetitions. Color is used to highlight the methods that yield the best results for each image. It can be observed that the proposed Hermite with 7-point FIR image derivatives achieves the highest PSNR in 16 out of 20 images (with oMOM deg 5 and Deep Learning LapSNR achieving best PSNR in 2 images. In terms of SSIM, the Hermite with IIR image derivatives achieved the best performance in 17 out of 20 images, with the oMOM deg 5 being the best method in 3 (out of 20) images and 1 image in tie between the two aforementioned methods.

\begin{landscape}
\begin{table}[htbp]
\centering
\caption{PSNR achieved by all methods under comparison for each image, after 20 repetitions (best values indicate in color).}
\label{psnr_table}
\setlength{\tabcolsep}{1pt}
\begin{tabular}{|m{1.7cm}|m{1.3cm}|m{1.3cm}|m{1.3cm}|m{1.3cm}|m{1.3cm}|m{1.3cm}|m{1.3cm}|m{1.3cm}|m{1.3cm}|m{1.3cm}|m{1.3cm}|m{1.3cm}|m{1.3cm}| m{1.3cm}|}
\hline
\rowcolor{gray!50}
 & \multicolumn{3}{c|}{\textbf{\footnotesize Kernel Based}} & \multicolumn{4}{c|}{\textbf{\footnotesize Generalized Convolution}} & \multicolumn{3}{c|}{\textbf{\footnotesize Deep Learning}} & \multicolumn{4}{c|}{\textbf{\footnotesize Hermite}} \\
\hline
\rowcolor{gray!50} \textbf{\footnotesize Image} & \textbf{\footnotesize Nearest} & \textbf{\footnotesize Bilinear} & \textbf{\footnotesize Bicubic} & \textbf{\footnotesize B-Spline 3} & \textbf{\footnotesize B-Spline 5} & \textbf{\footnotesize oMOM 4} & \textbf{\footnotesize oMOM 5} & \textbf{\footnotesize LAPSNR} & \textbf{\footnotesize ESPCN} &\textbf{\footnotesize FSRCNN} & \textbf{\footnotesize  FIR 7} & \textbf{\footnotesize  FIR 5} & \textbf{\footnotesize  FIR 3} & \textbf{\footnotesize  IIR}\\
\hline
\rowcolor{lavender} airplane & 13.9789 & 14.5733 & 16.2826 & 20.7689 & 20.7242 & 20.7450 & 20.6338 & 20.7452 & 19.9487 & 20.3019 & \cellcolor{yellow}  20.8902 & 20.4785 & 20.4138 & 20.3029 \\
arctichare & 16.1653 & 16.5568 & 17.8573 & 20.3521 & 20.2744 & 20.2902 & 20.2138 &\cellcolor{yellow} 21.6732 & 20.3223& 20.8963 &  20.3849 & 19.9625 & 19.9483 & 19.8919 \\
\rowcolor{lavender} baboon & 15.6938 & 16.0674 & 16.8692 & 19.6032 & 19.7959 & 19.7844 & 19.8090 & 19.5492 & 19.1706 & 19.3805 & \cellcolor{yellow} 19.8364 & 19.5891 & 19.4487 & 19.7227 \\
barbara & 14.6830 & 15.1143 & 17.7198 & 21.4615 & 21.5012 & 21.5031 & 21.4852 & 20.8910& 20.5493& 20.8196 & \cellcolor{yellow} 21.5577 & 21.3387 & 21.2847 & 21.3589 \\
\rowcolor{lavender} boat & 15.8791 & 16.3873 & 18.2344 & 23.0309 & 23.1686 & 23.1664 & 23.1520 & 21.3695 & 20.6271 & 21.2596 & \cellcolor{yellow} 23.2581 & 22.8891 & 22.7563 & 22.9847 \\
cat & 14.9593 & 15.5389 & 16.4353 & 22.1417 & 22.2904 & 22.2886 & 22.2753 & 19.2665 & 19.0057 & 19.1999 & \cellcolor{yellow} 22.4049 & 21.9933 & 21.8461 & 22.1147 \\
\rowcolor{lavender} fruits & 16.9229 & 17.3626 & 18.9296 & 22.0392 & 22.0816 & 22.0861 & 22.0532 &\cellcolor{yellow} 22.5253 & 21.6861 & 22.1928 & 22.1818 & 21.7774 & 21.7026 & 21.8262 \\
frymire & 10.3165 & 10.8214 & 12.0235 & 16.8943 & 17.2887 & 17.2556 & \cellcolor{yellow} 17.3614 & 14.1494 & 14.1640 & 14.2916 & 17.2737 & 17.0492 & 16.8125 & 17.3562 \\
\rowcolor{lavender} girl & 16.3423 & 16.6693 & 19.5499 & 23.9752 & 24.0209 & 24.0221 & 24.0039 & 23.2041 & 22.6643 & 23.1637 & \cellcolor{yellow} 24.0732 & 23.6959 & 23.6415 & 23.5875 \\
goldhill & 17.0245 & 17.6263 & 18.8161 & 23.6368 & 23.6799 & 23.6879 & 23.6381 & 23.4340 & 22.7654 & 23.0888 & \cellcolor{yellow} 23.8308 & 23.4709 & 23.3702 & 23.5270 \\
\rowcolor{lavender} lena & 15.5165 & 15.9859 & 18.7920 & 23.9165 & 23.9649 & 23.9707 & 23.9273 & 22.7220 & 21.9453& 22.4613 & \cellcolor{yellow} 24.0607 & 23.6572 & 23.5728 & 23.6590 \\
monarch & 15.4562 & 16.1737 & 16.3694 & 24.2650 & 24.4881 & 24.4764 & 24.5002 & 20.6106 & 20.3271 & 20.6611 & \cellcolor{yellow} 24.5432 & 24.1614 & 23.9954 & 24.3238 \\
\rowcolor{lavender} mountain & 11.5181 & 12.2248 & 13.6800 & 17.6614 & 17.7525 & 17.7580 & 17.7084 & 15.8753& 15.7709 & 15.8742 & \cellcolor{yellow} 17.8573 & 17.6761 & 17.5627 & 17.4953 \\
peppers & 13.8670 & 14.4778 & 17.5851 & 23.4312 & 23.3051 & 23.3410 & 23.1631 & 21.9124 & 21.3532 & 21.7376 & \cellcolor{yellow} 23.5345 & 23.1509 & 23.1009 & 22.6973 \\
\rowcolor{lavender} pool & 20.1182 & 20.6315 & 24.3089 & 30.9274 & 31.1676 & 31.1547 & 31.1818 & 27.0699 & 27.0729 & 26.9579& \cellcolor{yellow}31.2178 & 30.8301 & 30.6582 & 30.9388 \\
sails & 16.0689 & 16.6202 & 17.6804 & 22.7821 & 23.0652 & 23.0464 & 23.0964 & 20.9101 &  20.4932& 20.7882 & \cellcolor{yellow} 23.0973 & 22.7621 & 22.5709 & 22.9584 \\
\rowcolor{lavender} serrano & 13.2700 & 13.9410 & 15.2754 & 21.7259 & 21.9337 & 21.9187 & \cellcolor{yellow} 21.9629 & 18.6987& 18.5713 & 18.8536 & 21.9400 & 21.6286 & 21.4992 & 21.5797 \\
tulips & 14.2728 & 14.7993 & 17.0047 & 24.2568 & 24.4460 & 24.4359 & 24.4586 &  21.7958 & 21.4981 & 22.0677 & \cellcolor{yellow} 24.4879 & 24.0866 & 23.9499 & 24.2048 \\
\rowcolor{lavender} watch & 19.0952 & 19.6250 & 20.5156 & 26.9449 & 27.1997 & 27.1837 & 27.2256 & 23.9174 & 23.4855 & 23.5796 & \cellcolor{yellow} 27.2251 & 26.8505 & 26.6781 & 27.0079 \\
zelda & 15.9276 & 16.3474 & 19.3039 & 24.2378 & 24.2499 & 24.2571 & 24.2140 & 26.6032 & 15.4213 & 25.8155 & 24.3450 & 23.9444 & 23.8865 & 23.9349 \\

\hline
\end{tabular}
\end{table}
\end{landscape}

%%%%%%%%%%%%%%%%%%%%%%%%%%%%%%%%%%%%%%%%%%%%%%%%%%%%%%%%%%%%%%%
%%%%%%%%%%%%%%%%%%% SSIM %%%%%%%%%%%%%%%%%%%%%%%%%%%%%%%%%%%%%%

\begin{landscape}
\begin{table}[!h]
\centering
\caption{SSIM achieved by all methods under comparison for each image, after 20 repetitions (best values indicate in color).}
\label{SSIM_table}
\setlength{\tabcolsep}{1pt}
\begin{tabular}{|m{2cm}|m{1.3cm}|m{1.3cm}|m{1.3cm}|m{1.3cm}|m{1.3cm}|m{1.3cm}|m{1.3cm}|m{1.3cm}|m{1.3cm}|m{1.3cm}|m{1.3cm}|m{1.3cm}|m{1.3cm}|}
\hline
\rowcolor{gray!50}
 & \multicolumn{2}{c|}{\textbf{\footnotesize Kernel Based}} & \multicolumn{4}{c|}{\textbf{\footnotesize Generalized Convolution}} & \multicolumn{3}{c|}{\textbf{\footnotesize Deep Learning}} & \multicolumn{4}{c|}{\textbf{\footnotesize Hermite}} \\
\hline
\rowcolor{gray!50} \textbf{\footnotesize Image} & \textbf{\footnotesize Bilinear} & \textbf{\footnotesize Bicubic} & \textbf{\footnotesize B-Spline 3} & \textbf{\footnotesize B-Spline 5} & \textbf{\footnotesize oMOM 4} & \textbf{\footnotesize oMOM 5} & \textbf{\footnotesize LAPSNR} & \textbf{\footnotesize ESPCN} &\textbf{\footnotesize FSRCNN} & \textbf{\footnotesize  FIR 7} & \textbf{\footnotesize  FIR 5} & \textbf{\footnotesize  FIR 3} & \textbf{\footnotesize  IIR}\\
\hline
\rowcolor{lavender} airplane & 0.4699 & 0.5754 & 0.9580 & 0.9654 & 0.9649 & 0.9661 & 0.6872 & 0.6775 & 0.6886 & 0.9658 & 0.9618 & 0.9564 & \cellcolor{yellow} 0.9669 \\
arctichare & 0.5852 & 0.6852 & 0.9741 & 0.9785 & 0.9783 & 0.9788 & 0.8031 & 0.7804 & 0.7985 & 0.9790 & 0.9762 & 0.9728 & \cellcolor{yellow} 0.9791 \\
\rowcolor{lavender} baboon & 0.2083 & 0.2861 & 0.8189 & 0.8552 & 0.8526 & 0.8607 & 0.2893 & 0.2889 & 0.2961 & 0.8547 & 0.8427 & 0.8210 & \cellcolor{yellow}  0.8643 \\
barbara & 0.2923 & 0.4816 & 0.8468 & 0.8578 & 0.8569 & 0.8597 & 0.5294 & 0.5273 & 0.5387 & 0.8572 & 0.8523 & 0.8456 & \cellcolor{yellow} 0.8604 \\
\rowcolor{lavender} boat & 0.3255 & 0.4389 & 0.9171 & 0.9342 & 0.9330 & 0.9368 & 0.5232 & 0.5202 & 0.5321 & 0.9336 & 0.9267 & 0.9157 & \cellcolor{yellow} 0.9380 \\
cat & 0.2320 & 0.3761 & 0.9470 & 0.9587 & 0.9579 & 0.9602 & 0.4857 & 0.4794 & 0.4989 & 0.9585 & 0.9540 & 0.9465 & \cellcolor{yellow} 0.9606 \\
\rowcolor{lavender} fruits & 0.5156 & 0.6257 & 0.9396 & 0.9495 & 0.9487 & 0.9511 & 0.6958 & 0.6857 & 0.6975 & 0.9491 & 0.9443 & 0.9379 & \cellcolor{yellow} 0.9518 \\
frymire & 0.1086 & 0.4033 & 0.9609 & 0.9710 & 0.9702 & 0.9728 & 0.3643 & 0.3569 & 0.3627 & 0.9705 & 0.9656 & 0.9594 & \cellcolor{yellow} 0.9733 \\
\rowcolor{lavender} girl & 0.6220 & 0.6892 & 0.9354 & 0.9457 & 0.9450 & 0.9474 & 0.7121 & 0.7027 & 0.7098 & 0.9455 & 0.9409 & 0.9343 & \cellcolor{yellow} 0.9479 \\
goldhill & 0.3034 & 0.4034 & 0.9211 & 0.9370 & 0.9359 & 0.9393 & 0.5373 & 0.5344 & 0.5419 & 0.9368 & 0.9307 & 0.9205 & \cellcolor{yellow} 0.9403 \\
\rowcolor{lavender} lena & 0.4116 & 0.6100 & 0.9532 & 0.9613 & 0.9608 & 0.9625 & 0.6811 & 0.6733 & 0.6856 & 0.9611 & 0.9572 & 0.9517 & \cellcolor{yellow} 0.9626 \\
monarch & 0.4758 & 0.5498 & 0.9671 & 0.9723 & 0.9719 & \cellcolor{yellow} 0.9731 & 0.7231 & 0.7189 & 0.7301 & 0.9721 & 0.9691 & 0.9656 & \cellcolor{yellow} 0.9731 \\
\rowcolor{lavender} mountain & 0.1993 & 0.2880 & 0.8380 & 0.8643 & 0.8624 & 0.8683 & 0.2770 & 0.2702 & 0.2790 & 0.8645 & 0.8547 & 0.8383 &\cellcolor{yellow}  0.8710 \\
peppers & 0.4467 & 0.6239 & 0.9604 & 0.9652 & 0.9648 & 0.9658 & 0.6838 & 0.6782 & 0.6930 & 0.9651 & 0.9619 & 0.9584 & \cellcolor{yellow} 0.9656 \\
\rowcolor{lavender} pool & 0.8076 & 0.8779 & 0.9588 & 0.9620 & 0.9617 & 0.9626 & 0.8987 & 0.8914 & 0.8942 & 0.9616 & 0.9585 & 0.9557 & \cellcolor{yellow} 0.9623 \\
sails & 0.1967 & 0.2845 & 0.8691 & 0.8967 & 0.8947 & 0.9010 & 0.3681 & 0.3672 & 0.3740 & 0.8959 & 0.8860 & 0.8690 &\cellcolor{yellow}  0.9036 \\
\rowcolor{lavender} serrano & 0.2520 & 0.3781 & 0.9408 & 0.9500 & 0.9493 & 0.9516 & 0.5472 & 0.5411 & 0.5556 & 0.9492 & 0.9442 & 0.9380 & \cellcolor{yellow} 0.9516 \\
tulips & 0.2354 & 0.4224 & 0.9460 & 0.9570 & 0.9562 & 0.9587 & 0.5855 & 0.5769 & 0.5992 & 0.9564 & 0.9512 & 0.9440 & \cellcolor{yellow} 0.9591 \\
\rowcolor{lavender} watch & 0.4952 & 0.5728 & 0.9777 & 0.9821 & 0.9818 & \cellcolor{yellow} 0.9828 & 0.7243 & 0.7209 & 0.7213 & 0.9819 & 0.9794 & 0.9764 & 0.9824 \\
zelda & 0.4670 & 0.6567 & 0.9680 & 0.9731 & 0.9728 & \cellcolor{yellow} 0.9738 & 0.7722 & 0.7615 & 0.7729 & 0.9730 & 0.9703 & 0.9670 & 0.9735 \\

\hline
\end{tabular}
\end{table}
\end{landscape}

\section{Conclusions and further work}
In this work we proposed the construction of Hermite kernels for 2D image interpolation, based on the Hermite splines that were defined and studied on $n-$dimensional grid \cite{DK23}. This formulation makes the Hermite splines easy to implement. In this work we also calculated the kernels for 2D image zooming and perform comparisons with many state-of-the-art methods for 20 images. The obtained results reveal several interesting aspects. The generalized convolution methods achieved very good results both in terms pf PSNR and SSIM. The convolution methods were inferior. The proposed Hermite kernels outperformed almost all methods in the majority of the 20 images, up to 20 repetitions of the zoom in and out operator. The Deep learning based methods exhibited an increasing blurring as the number of repetitions increased. However the resulting PSNR appeared disproportionally high. This is due to small MSE (mean square error), due to small errors in relatively flat areas, despite the apparent loss of spatial resolution in areas with fine detail. The SSIM error metric was low, since, as expected, it was sensitive to this issue. 
Future work will investigate similar tasks in 3d images or videos.

%\section{Error of interpolation}
%\section{Remainder of the interpolation for $\Bbbk=\mathbb{R}$ or $\mathbb{C}$}
%\label{errr}
%\input{error}

%\section{On the ideal of the interpolation}
%\label{idealco}
%\input{onideal_new}
%Previous Section
%\input{idealconn}

%\newpage
%\section{An alternative quick proof of Theorem \ref{th1}}
%\label{altproof}
%\input{newdef}
%\newpage

%\section{Computational Complexity}
%\label{comp}
%\input{rema1}

%\newpage
%\section{Examples}
%\label{examm}
%\input{examp}

%\newpage
%\section{Conclusions}
%\label{conc}
%\input{conc}


%\bibliographystyle{elsarticle-num}
%\bibliography{refs}
%\nocite{*}


\end{document}